# A Study of AI Population Dynamics with Million-agent Reinforcement Learning

Yaodong Yang[†], Lantao Yu[‡], Yiwei Bai[‡], Jun Wang[†], Weinan Zhang[‡], Ying Wen[†], Yong Yu[‡]

University College London[†], Shanghai Jiao Tong University[‡]

## ABSTRACT

We conduct an empirical study on discovering the ordered collective dynamics obtained by a population of intelligence agents, driven by million-agent reinforcement learning. Our intention is to put intelligent agents into a simulated natural context and verify if the principles developed in the real world could also be used in understanding an artificially-created intelligent population. To achieve this, we simulate a large-scale predator-prey world, where the laws of the world are designed by only the findings or logical equivalence that have been discovered in nature. We endow the agents with the intelligence based on deep reinforcement learning (DRL). In order to scale the population size up to millions agents, a large-scale DRL training platform with redesigned experience buffer is proposed. Our results show that the population dynamics of AI agents, driven only by each agent's individual self-interest, reveals an ordered pattern that is similar to the *Lotka-Volterra* model studied in population biology. We further discover the emergent behaviors of collective adaptations in studying how the agents' grouping behaviors will change with the environmental resources. Both of the two findings could be explained by the *self-organization* theory in nature.

## KEYWORDS

Multi-agent reinforcement learning, deep Q-learning, population dynamics, self-organization theory, collective behaviors, grouping

## 1 INTRODUCTION

By employing the modeling power of deep learning, single-agent reinforcement learning (RL) has started to display, even surpass, human-level intelligence on a wide variety of tasks, ranging from playing the games of Labyrinth [31], Atari [33], and Go [43] to other tasks such as continuous control on locomotions [27], text generation [53], and neural architecture design [54]. Very recently, multi-agent RL algorithms have further broadened the use of RL and demonstrated their potentials in the setting where both of the agents' incentives and economical constraints exist. For example, the studies [8, 25, 34, 49] have shown that with different multi-agent cooperative learning environments, the compositional language naturally emerges. Researchers in [13, 40, 48] have also demonstrated that multiple agents can be trained to play the combat game in StarCraft, and the agents have mastered collaborative strategies that are similar to those of experienced human players. Nonetheless, all of the aforementioned RL systems so far have been limited to less than tens of agents, and the focuses of their studies are rather in the optimization of a micro and individual level policy. Macro-level studies about the resulting collective behaviors and dynamics emerging from a large population of AI agents remain untouched.

Yet, on the other hand, real-world populations exhibit certain orders and regularity on collective behaviors: honey bees use specific waggle dance to transmit signals, a trail of ants transfer food by leaving chemical marks on the routes, V-shaped formations of bird flocks during migration, or particular sizes of fish schools in the deep ocean. Even human beings can easily show ordered macro dynamics, for example, the rhythmical audience applause after the concerts, the periodical human waves in the fanatic football game, *etc*. A stream of research on the theory of *self-organization* [2] explores a new approach to explaining the emergence of orders in nature. In fact, the self-organizing dynamics appears in many other disciplines of natural sciences [3]. The theory of *self-organization* suggests that the ordered global dynamics, no matter how complex, are induced from repeated interactions between local individual parts of a system that are initially disordered, without external supervisions or interventions. The concept has proven important in multiple fields in nature sciences [7, 23, 45].

As once the ancient philosopher *Lucretius* said: "*A designing intelligence is necessary to create orders in nature.*" [38], an interesting question for us is to understand what kinds of ordered macro dynamics, if any, that a community of artificially-created agents would possess when they are together put into the natural context. In this paper, we fill the research gap by conducting an empirical study on the above questions. We aim to understand whether the principles, *e.g.*, self-organization theory [2], that are developed in the real world could also be applied on understanding an AI population. In order to achieve these, we argue that the key to this study is to have a clear methodology of introducing the micro-level intelligence; therefore, we simulate a predator-prey world where each individual AI agent is endowed with intelligence through a large-scale deep reinforcement learning framework. The population size is scaled up to million level. To maximize the generality, the laws of the predator-prey world are designed by only incorporating the natural findings or logic equivalence; miscellaneous potential dynamics can thus be studied. We first study the macro dynamics of the population size for both the predators and preys, and then we investigate the emergence of one most fundamental collective behavior – grouping. In particular, we compare the statistics and

Yaodong Yang[†], Lantao Yu[‡], Yiwei Bai[‡], Jun Wang[†], Weinan Zhang[‡], Ying Wen[†], Yong Yu[‡]
University College London[†], Shanghai Jiao Tong University[‡]

dynamics of the intelligent population with the theories and models from the real-world biological studies. Interestingly, we find that the artificial predator-prey ecosystem, with individual intelligence incorporated, reaches an ordered pattern on dynamics that is similar to what the *Lotka-Volterra* model indicates in population biology. Also, we discover the emergence of the collective adaptations on grouping behaviors when the environment changes. Both of the two findings can be well explained based on the *self-organization* theory. Moreover, the proposed million-agent RL platform could serve as the initial steps to understand the behaviors of large-scale AI population, driven by deep reinforcement learning. It could potentially open up an interesting research direction of understanding AI population by linking the findings from the AI world with the natural science principles from the real world, thus contributing to the research frontiers such as smart city [35] and swarm intelligence [24].

## 2 RELATED WORK
### 2.1 Reinforcement learning

Reinforcement learning (RL) [46] employs a goal-oriented learning scheme that reinforces an agent to maximize its cumulative rewards through sequentially interacting with the environment. The intelligence evolves by agent's learning from the past experiences and trying to perform better in the future. Recently, deep neural networks succeed in marrying the RL algorithms; in particular, they show remarkable performance in approximating the value function [32], the policy function [27], or both the value and policy function (*a.k.a.* actor-critic) [31], all of which increase the "intelligence" of traditional RL methods. Single-agent RL methods have been extended to the multi-agent settings where multiple agents exist and interact with each other. Q-learning methods such as minimax Q-learning [19], Nash Q-learning [18] have been proposed.

In addition to the work where minimal or even no communication between learning agents are considered [9, 10, 30], a fundamental question to answer in multi-agent RL is how different agents should communicate so as to reach a coherent goal. Several differentiable communication protocols have been proposed [12, 44], which can be easily embedded into the error back-propagation training scheme. The work in [40] employed bidirectional recurrent neural networks to coordinate groups of agents to play StarCraft combat games, and achieved human-level micro-management skills. Beyond pursuing high performance on playing video games, researchers recently start to shift the focus onto studying the community of AI agents, and its corresponding attributes. A few concurrent studies [8, 25, 34, 49] were conducted in different cooperative learning environments, through which the emergence of compositional language has been found. Leibo et al. introduced agents' self-interested policy learning into solving sequential social dilemmas, and discovered how agents' behaviors would be influenced by the environmental factors, and when conflicts would emerge from competing over shared resources. Nonetheless, the multi-agent systems in those studies consider no more than tens of agents; it is thus unfair to generalize the findings to the population level. Macro dynamics of large AI population remain to be disclosed.

### 2.2 Population Biology

While our subject is computerized artifacts, our work is also related to the research conducted in natural sciences. The theory of *self-organization* proposed in [2] serves as a fundamental way of thinking to understand the emergence of orders in nature (even though Physicists tend to challenge this theory because *The Second Law of Thermodynamics* [5] states that the total level of disorders in an isolated system can never decrease over time). *Self-organization* theory believers think that the global ordered dynamics of a system can originate from numerous interactions between local individuals that are initially disordered, with no needs for external interventions. The theory predicts the existence of the ordered dynamics in the population. In fact, the self-organizing phenomena have been observed in multiple fields in natural sciences [7, 23, 45]. For example, in population biology, one important discovery is the ordered harmonic dynamics of the population sizes between predators and preys (*e.g.*, the lynx and snowshoe hare [15]), which is summarized into the *Lotka-Volterra* model [28]. It basically describes the fact that there is 90° lag in the phase space between the population sizes of predators and preys (more details are discussed later in Section 5.1). Even though explainable via the *self-organization* theory, the *Lotka-Volterra* models are summarized based on the statistics from the ecological field studies. There is essentially no learning process or individual intelligence involved. In this work, we chose a different approach by incorporating the individual intelligence into the population dynamics studies where each agent is endowed with the intelligence to make its own decision rather than is considered as homogeneous and rule-based. Our intention is to find out whether an AI population still creates ordered dynamics such as the *Lotka-Volterra* equations, if so, whether the dynamics is explainable from the perspective of the *self-organization* theory.

In multiple disciplines of natural sciences spanning from zoology, psycholog, to economy [11, 17, 45], one of the most fundamental thus important collective behaviors to study is: grouping – a population of units aggregate together for collective decision-making. Grouping is believed to imply the emergence of sociality and to induce other collective behaviors [21]. In studying the grouping behaviors, traditional approaches include setting up a game with rigid and reductive predefined interactive rules for each agent and then conduct simulations based on the ad-hoc game [14, 20, 37]. Rule-based games might work well on biological organisms that inherit the same characteristics from their ancestors; however, they show limits on studying the large-scale heterogeneous agents [4]. In contrast to rule-based games with no learning process involved, here we investigate the formation of grouping behaviors on a million-level AI population driven by RL algorithms. Our intention is to find out how the grouping behaviors in AI population emerge and change *w.r.t.* the environmental factors such as the food resources, and if there is any other collective behaviors emerging from grouping.

## 3 DESIGN OF THE LARGE-SCALE PREDATOR-PREY WORLD

In this paper, we try to understand: 1) whether AI population create any ordered patterns on population dynamics, and 2) the dynamics of the collective grouping behaviors. Predator-prey interaction is one fundamental relationship observed in nature. Here we intend to simulate a predator-prey world with million-level agents (shown in Fig. 1). The world is deigned to be easily adaptable to incorporate other environmental complexity to investigate the miscellaneous dynamics as well as collective behaviors of AI population where each individual agent is driven by purely self-interest.

Table 1: Natural Evidence of the Axioms in the Simulated Predator-prey Environment

| Axiom | Examples in nature |
|---|---|
| Positive Feedback | The observations on ants [6, 52] suggest that when an ant discovers a food source through a particular search trail, the path will soon serve as the trigger for a positive feedback, by its leaving chemical landmarks, through which other ants start to follow. |
| Negative Feedback | In the ant's case, as the population size of ants is limited, with increasing number of ants forage the food from outside, the distribution of ants between food sources will be stable [16]. |
| Individual Variation | Social insects as honey bees are evolved to be highly variable in the directional sense, response to sucrose, level of focus in food collection in order to ensure the diversification in ways of food collection, otherwise one single food resource will be depleted quickly [22, 39]. |
| Response Threshold | Bumble bees will start to fan so as to cool down the hive when the temperature inside goes above a threshold level [51]. |
| Redundancy | In the kingdom of bees, if the community suffers a drastic reduction in the number of worker bees, younger bees will soon replace their positions to guarantee that the whole community function well [42]. |
| Synchronisation | In a concert, individuals with unique frequency of applause could affect the frequency of the crowd through implicit synchronisation [36]. Empirically, audience applause are often achieved through adjustments by individuals having an unique frequency among the local average. |
| Selfishness | Easily observable in nature. A typical example would be the Praying Mantis female who eats the male head after mating as a reproductive strategy to enhance fertilization [47]. |

## 3.1 The Axioms of Natural Environments

To avoid introducing any specific rules that could harm the generality of the observed results, we design the laws of the world by only considering those real findings or logical equivalence that have been observed in the natural system. We regard those laws as the *axioms* of studying population dynamics and collective behaviors. Here we briefly review the axioms accepted, and refer the corresponding natural evidence to the Table.1. Note that these axioms should not be treated separately, we consider instead how the combination of these different axioms could produce and affect collective dynamics.

(1) *Positive Feedback*. Positive feedback enhances particular behaviors through reinforcement. It helps spread the information of a meaningful action quickly between individuals.
(2) *Negative Feedback*. Negative feedback leads to homeostasis. It helps stabilize the collective behaviors produced in favor of the positive feedback from going to extremes.
(3) *Individual Variation*. Individual variation is of the essence to guarantee the continual explorations of new solutions to the same problem within a population.
(4) *Response Threshold*. Response threshold is the threshold beyond which individuals will change their behaviors as a response to the stimulus.
(5) *Redundancy*. Redundancy ensures functional continuity of the whole population even when a catastrophic event happens.
(6) *Synchronization*. Synchronization is a special kind of positive feedback in time rather than space. An example would be how individual with unique frequency of applause affect the frequency of the crowd in a concert.
(7) *Selfishness*. Individuals always tend to maximize their own utility. One will not behave altruistically for others until he can benefit more from behaving collectively than acting alone.

## 3.2 Realization of the Axioms

We realize the predator-prey world via designing a *Stochastic Game*. We list the detailed rules of the game and its corresponding axiom.

*3.2.1 Population Dynamics.* In the predator-prey world (see Fig. 1), the goal for the predator species is to survive in the ecosystem and procreate their next generations (Axiom.7). Positions of predator/prey/obstacles in the world are all initialized randomly at the beginning. The environment is considered under an infinite horizon. While the population of both preys and predators can be boomed by breeding offsprings (Axiom.5), they however face the hazards of either being hunted as preys, or dying of starvation as predators (Axiom.2). To realize the idea of starvation, we make the health status of predator decrease with time by a constant factor, which can also be restored by capturing and eating preys (Axiom.1). Predators are assumed to have infinite appetite; the logical equivalence in nature is that a predator normally has the ability of storing food resource for future survival. Each predator can have unique characteristics, *e.g.*, identity vector, eyesight and health status (Axiom.3). The unique characteristics of each agent represents the diversity of the population. Each individual agent make independent decision, and can behave differently even given the same scenario. Predators can form a group to increase the chance of capturing a prey (Axiom.1,4). Group members are visible to predators within its view. If a single agent chooses the action of "join a group", the environment will select a group within its view randomly, and the agent will become a member of that group until it decides to "leave the current group" afterwards. Note that a single predator may hunt for the prey alone as well as hunt as a group member. As illustrated in Fig. 1, each prey is assigned a square capture area with a capture radius $\rho$, which reflects the difficulty of being hunted (Axiom.4). Groups of predators, or singles, will only be able to hunt the prey if they manage to stay within the capture radius. Apart from the capture radius, another parameter,

AAMAS'18, July 2018, Stockholm, Sweden

Yaodong Yang[†], Lantao Yu[‡], Yiwei Bai[‡], Jun Wang[†], Weinan Zhang[‡], Ying Wen[†], Yong Yu[‡]
University College London[†], Shanghai Jiao Tong University[‡]

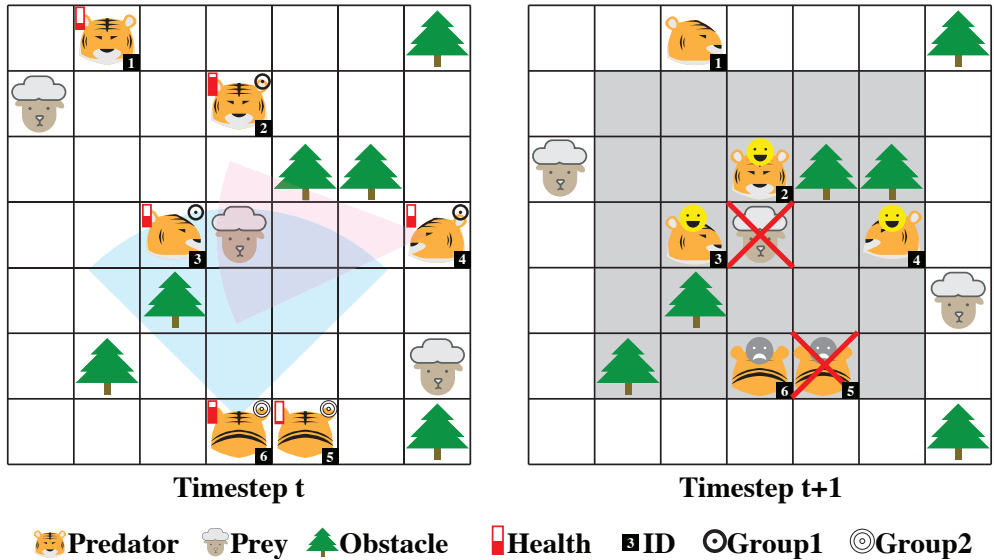

**Figure 1: Illustration of the predator-prey world.** In the 2-D world, there exist preys, predators, and obstacles. Predators hunt the prey so as to survive from starvation. Each predator has its own health bar and limited eyesight view. Predators can form a group to hunt the prey so that the chance of capturing can increase, but this also means that the captured prey will be shared among all group members. When there are multiple group targeting the same prey, the largest group within capture radius will win. In this example, predators $\{2, 3, 4\}$ form a group and win the prey over the group $\{5, 6\}$. Predator 5 soon dies because of starvation.

the capture threshold $k$ ($k$=0,1,2,...), also reflects the capturing difficulty of each prey (Axiom.4). Within the capture area, only meeting the threshold will a group of predators become a valid candidate. When there are multiple valid candidate groups targeting at the same prey, the group with the largest group size will be the winner, which mimics the law of jungle. When a group wins over other candidates, all the members in that group will share the prey equally (Axiom.2). The trade-off here is, in the pursuit of preys, grouping is encouraged as large group can help increase the probability of capturing a prey; however, huge group size will also be inhibited due to the less proportion of prey each group member obtains from the sharing (Axiom.7).

*3.2.2 Grouping Behaviors.* Considering *synchronization* of Axiom.6 and *selfishness* of Axiom.7, we incorporate a second type of prey that can be captured by an individual predator alone, which means we set the capture threshold $k$ to 1 for that species. An analogy here is to think of tigers as the predators, sheep as the preys whose captures require collaborative grouping between predators, and rabbits as the preys that can be captured by a single predator. These two kinds of preys can be considered as an abstraction of individual reward and grouping reward respectively. Predators have to make a decision to either join a group for hunting the sheep or conduct hunting the rabbit by itself in order to maximize its long-term reward and the probability of survival (Axiom.1,2,7), which introduces a trade-off between acting alone and collaborating with others. We keep alternating the environments by feeding these two kinds of preys one after another (Axiom.6) and examine the dynamics of grouping behaviors. To emphasize the dynamics of grouping behaviors and also to avoid the influences from the systematic preferences for grouping as a result of the changing population size, we keep the population size of predators fixed by endowing them with eternal longevity, which can also be considered as a short-term observation during which there is little change of the predator population size. Under the environment, the optimal strategy for each agent continuously varies over time and the predator population has to learn to adapt their collective strategy correspondingly.

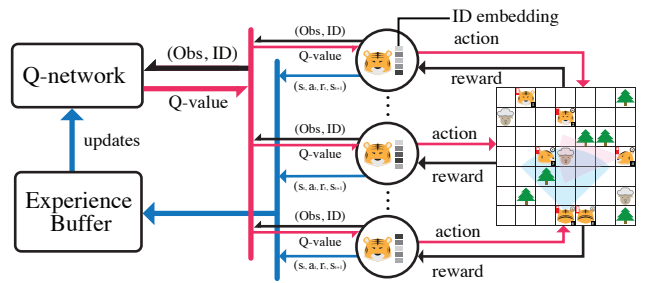

**Figure 2: Million-agent Q-learning System in the Predator-prey World.**

## 4 AI POPULATION BUILT BY MULTI-AGENT DEEP REINFORCEMENT LEARNING

### 4.1 Multi-agent Markov Decision Process

In the designed predator-prey world, we build AI population under the multi-agent deep reinforcement learning setting. Formally,

**Algorithm 1** Million-agent Q-learning (in the case of population dynamics in Section 3.2)

1: Initialize agent's Q-network $\pi^i$, agent's identity $v^i$.
2: Randomly initialize the environment $s \sim \rho_0(\mathcal{S})$.
3: **for** time step=1,2,..., **do**
4:     Procreate predators/preys offsprings with random positions.
5:     **for** agent i=1,2,...,n **do**
6:         Compute the local observation features $O(i)$.
7:         Compute the identity embedding $\mathcal{I}(i)$.
8:         Compute the current state for agent: $s_t^i = (O(i), \mathcal{I}(i))$.
9:         Take action $a_t^i \sim \pi_\theta^i(a^i|s^i) = \epsilon\text{-greedy}(Q^{\pi^i}(s_t^i, a_t^i))$.
10:        Apply action $a_t^i$, and get reward $r_t^i$, next state $s_{t+1}^i$.
           *Within the capture radius of each prey, the group of predators meeting the threshold will become valid candidate, the candidate with largest group size will be the final winer, and the reward are shared among the group members equally.*
11:        Store tuple $< s_t^i, a_t^i, s_{t+1}^i, r_t^i >$ in the experience buffer.
12:     **end for**
13:     **while** $|\mathcal{B}| \geq$ batch size **do**
14:         Sample a mini-batch from $\mathcal{B}$.
15:         Update the parameters of Q-function *w.r.t.* the loss:
16:            $(r_t^i + \gamma \max_{a' \in \mathcal{A}} Q^\pi(s_{t+1}^i, a') - Q^\pi(s_t^i, a_t^i))^2$
17:     **end while**
18:     Clear experience buffer $\mathcal{B}$.
19:     Decay the health of predators who starve.
20:     Reward (the group of) predator(s) who win the preys.
21:     Remove the dead predators and preys from the map.
22: **end for**

the multi-agent Markov decision process (or, stochastic game) is denoted by $\{\mathcal{S}, \mathcal{A}, \mathcal{T}, \mathcal{R}, O, \gamma, \rho_0, N\}$. $\mathcal{S}$ denotes the set of true environmental states, and $\rho_0(\mathcal{S})$ denotes the initial state distribution. At each time step, each agent $i \in \{1, ..., N\}$ in the predators population (they have to hunt preys to survive) takes an action $a^i \in \mathcal{A}$ where $\mathcal{A}$ is the valid action space. The joint actions $\mathbf{a} \in \mathcal{A}^N$ induce a transition of the environment based on the transition function between states, $\mathcal{T} : \mathcal{S} \times \mathcal{A}^N \to \mathcal{S}$. The reward function is defined by $\mathcal{R} : \mathcal{S} \times \mathcal{A}^N \to \mathbb{R}^N$, and $\gamma \in [0, 1)$ denotes the discount factor. The environment is partially-observed; each agent can observe $o^i \in O(s, a^i)$. An agent gains "intelligence" by learning a stochastic policy $\pi_\theta^i(a^i|s^i = o^i)$ that could maximize its expected cumulative reward in the predator-prey environment, *i.e.*, $\theta^* := \arg\max_\theta \mathbb{E}_{(s,\mathbf{a})}[\sum_{t=0}^\infty \gamma^t \mathcal{R}_t^i]$. The action-value function is defined by $Q^{\pi^i}(s_t, a_t) = \mathbb{E}_{s_{t+1:\infty}, a_{t+1:\infty}}[\sum_{l=0}^\infty \gamma^l \mathcal{R}_{t+l}^i | s_t, a_t]$. Considering the exploration in the action space, $\epsilon$-greedy methods can be applied on selecting the action, $\pi_\theta^i(a^i|s^i) = \epsilon\text{-greedy}(Q^{\pi^i}(s^i, a^i))$.

The action space $\mathcal{A}$ includes {forward, backward, left, right, rotate left, rotate right, stand still, join a group, and leave a group}. It is considered as invalid if a predator takes the "join a group" action as a group member already, takes the "leave a group" action as a single individual, or tries to cross the map boarders. Invalid actions will not be settled by the environment. Within the horizon of each individual agent, there are five channels for the observation $o^i$. The observation $O_t^i \in \mathbb{R}^{m \times n \times 5}$ is dependent on the agent's current position and orientation. The agent's eyesight ranges up to a distance limit towards the grids ahead and the grids to the left and right. The type of object (predators/preys/obstacles/blank areas) on the map occupy the first three channels, which are the raw RGB pixels. The fourth channel is an indicator of whether or not that object is a group member. The fifth channel is the health status $h \in \mathbb{R}$ if the object is an agent, otherwise padded with zero. Each agent is assigned with an unique identity embedding $v^i \in \mathbb{R}^5$, together with the local observation, it makes up the state for each agent $s^i = (o^i, v^i)$. Individual agent is supposed to make independent decisions, and behave differently based on its local observation as well as ID embeddings as the inputs of policy $\pi_\theta^i$.

### 4.2 The Implementation

We designed a multi-agent reinforcement learning platform with environmental optimizations in TensorFlow [1] to make the training of million-agent learning feasible. To the best of our knowledge, we are the first[1] to introduce the training environment that enables simulating millions of agents driven by deep reinforcement learning algorithms. The demonstration of the platform will be presented during NIPS 2017, for double blind review, we omit the author details here.

In particular, our setting is implemented through "centralised training with independent execution". This is a natural paradigm for a large set of computationally tractable multi-agent problems. In the training stage, agents update the centralised Q-value function approximated by a deep neural network: $Q^\pi(s^i, a^i) = Q((o^i, v^i), a^i)$. Each individual agent, however, must rely on its local observation as well as unique identity to make independent decisions during the execution time. Apart from the standard setting of Q-learning [50] and deep Q-learning [32], here we introduce a special experience buffer consdiering the GPU efficiency as well as mitigating the non-stationary issue in the off-policy learning. At each time step, all agents contribute its experienced transitions $(s_t^i, a_t^i, r_t^i, s_{t+1}^i)$ to the buffer, as shown in Fig. 2. We collect all the agents' experience of one time step in parallel and then update the Q-network using the experience at the same time. This significantly increases the utilization of the GPU memory, and is essential to the million-agent training. Based on the experience from the buffer, the Q-network is updated as:

$$Q^\pi(s_t^i, a_t^i) \leftarrow Q^\pi(s_t^i, a_t^i) + \alpha[r_t^i + \gamma \max_{a' \in \mathcal{A}} Q^\pi(s_{t+1}^i, a') - Q^\pi(s_t^i, a_t^i)]. \quad (1)$$

It is worth mentioning that the experience buffer in Fig. 2 stores the experience from the agents only for the current time step; this is markedly different from the replay buffer that is commonly used in the traditional DQN where the buffer maintains a first-in-first-out queue across different time steps. Using the off-policy replay buffer will typically lead to the non-stationarity issue for the multi-agent learning tasks [29]. On the other hand, Mnih et al. introduced the replay buffer aiming at disrupting the auto-correlations between the consecutive examples. In our million-agent RL setting, the experiences are sampled concurrently from millions of agents, each individual agent with different states and policies; therefore, there is naturally no strong auto-correlations between the training examples. Moreover, it is unlikely that the unwanted feedback loops

---
[1]The Github address of the platform will be presented in the final version of this paper.

Yaodong Yang[†], Lantao Yu[‡], Yiwei Bai[‡], Jun Wang[†], Weinan Zhang[‡], Ying Wen[†], Yong Yu[‡]
University College London[†], Shanghai Jiao Tong University[‡]

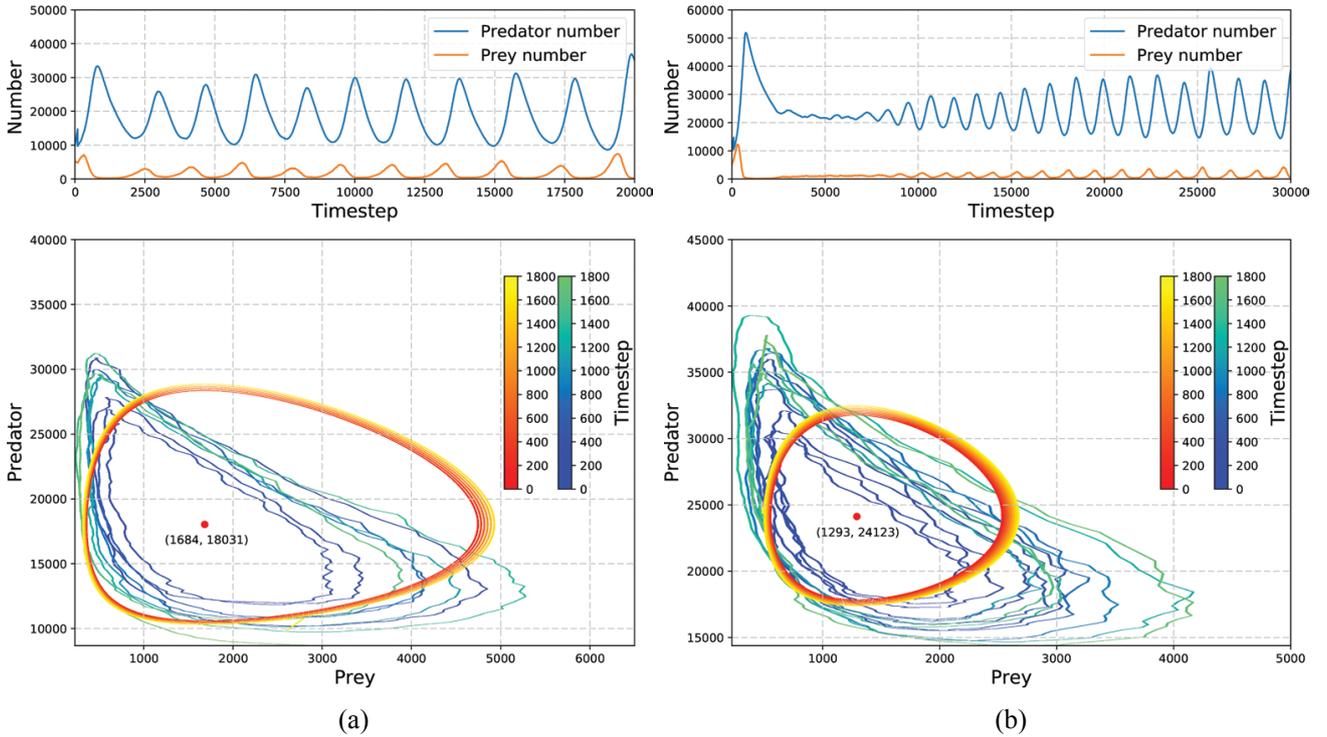

Figure 3: Population dynamics in both the time space ($1^{st}$ row) and the phase space ($2^{nd}$ row). The orange circles denote the theoretical solutions to the *Lotka-Volterra* equation, with the red spot as the equilibrium. The green-blue circles denote the simulation results. a): The simulated birth rate of preys is 0.006. Fitted LV model: $\alpha = 0.0067, \beta = 3.75 \times 10^{-7}, \delta = 6.11 \times 10^{-7}, \gamma = 0.001$. b): The simulated birth rate of preys is 0.01. Fitted LV model: $\alpha = 0.0086, \beta = 3.57 \times 10^{-7}, \delta = 9.47 \times 10^{-7}, \gamma = 0.0012$, where $\alpha$ in the LV model represents the birth rate.

arise since the sampled experiences will hardly be dominated by one single agent's decision. The results further testify the robustness of our design of the experience buffer. See Algorithm.1 for the pesudocode of the population dynamics example described in Section 3.2.

## 5 EXPERIMENTS AND FINDINGS

Two sets of experiments – understanding population dynamics & collective behaviors – have been conducted. The environmental parameter settings (*e.g.*, the eyesight limit of predators), and the code to reproduce the results with no needs for further adjustments will be released in *Supplementary Material* in the final version.

### 5.1 Understanding the Population Dynamics

We first study the population dynamics with a community of predators and preys by tracking the population size of each species over time. Specifically, we initialize 10,000 predators and 5,000 preys randomly scattered over a map of size $1,000 \times 1,000$. All predators' health status is set to 1.0 initially and decays by 0.01 at each time step. In two comparing settings, the birth rates of preys are set to 0.006 and 0.01 respectively. The Q-network has two hidden layers, each with 32 hidden units, interleaved with sigmoid non-linear layers, which then project to 9-dimensional outputs, one for each potential action. During training, the predators learn in an off-policy reinforcement learning scheme, with exploratory parameter $\epsilon = 0.1$.

Surprisingly, we find that the AI population reveals an ordered pattern when measuring the population dynamics. As shown in Fig. 3, the population sizes of both predators and preys reach a dynamic equilibrium where both curves present a wax-and-wane shape, but with a 90° lag in the phase, *i.e.*, the crest of one is aligned with the trough of the other. The underlying logic of such ordered dynamics could be that when the predators' population grows because they learn to know how to hunt efficiently, as a consequence of more preys being captured, the preys' population shrinks, which will later cause the predators' population also shrinks due to the lack of food supply, and with the help of less predators, the population of preys will recover from the shrinkage and start to regrow. Such logic drives the 2-D contour of population sizes (see the green-blue traits in the $2^{nd}$ row in Fig. 3) into harmonic cycles, and the circle patterns become stable with the increasing level of intelligence agents acquire from the reinforcement learning. As it will be shown later in the ablation study, enabling the individual intelligence is the key to observe these ordered patterns in the population dynamics.

In fact, the population dynamics possessed by AI agents are consistent with the *Lotka-Volterra* (LV) model studied in biology (shown by the orange traits in Fig. 3). In population biology, the LV model [28] describes a *Hamiltonian* system with two-species interactions, *e.g.*, predators and preys. In the LV model, the population size of predators $q$ and of preys $p$ change over time based on the following pair of nonlinear differential equations:

$$\frac{1}{p}\frac{dp}{dt} = \alpha - \beta q, \qquad \frac{1}{q}\frac{dq}{dt} = \delta p - \gamma. \qquad (2)$$

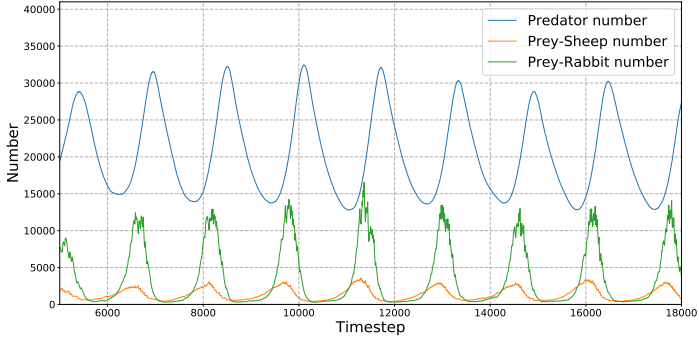

Figure 4: Population dynamics in the time space and the phase space. A new type of prey (green line) is introduced, which can be captured by a single agent. The AI population shows ordered dynamics in the 3-D phase space.

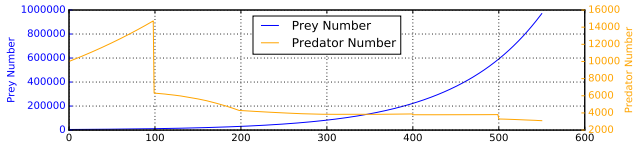

Figure 5: Population dynamics with the learning function of AI population disabled. The simulation follows the same setting as Fig. 3(b). No ordered dynamics are found any more.

The preys are assumed to have an affluent food resource and thus can reproduce exponentially with rate $\alpha$, until meeting predation, which is proportional to the rate at which the predators and the prey meet, represented by $\beta q$. The predators have an exponential decay in the population due to natural death denoted by $\gamma$. Meanwhile, they can also boost the population by hunting the prey, represented by $\delta p$. The solution to the equations is a harmonic function (wax-and-wane shaped) with the population size of predators lagging that of preys by $90°$ in the phase. On the phase space plot, it shows as a series of periodical circle $V = -\delta p + \gamma \ln(p) - \beta q + \alpha \ln(q)$, with $V$ dependent on initial conditions. In other words, which equilibrium cycle to reach depends on where the ecosystem starts. Similar patterns on the population dynamics might indicate that the orders from an AI population is induced from the same logic as the ecosystem that LV model describes. However, the key difference here is that, unlike the LV equations that model the observed macro dynamics directly, we start from a microcosmic point of view – the AI population is only driven by the self-interest (powered by RL) of individual agent, and then reaching the macroscopic principles.

To further test the robustness of our findings, we perform an ablation study on three of the most important factors that we think of are critical to the generation of the ordered dynamics. First, we analyze whether the observed pattern is restricted by the specific settings of the predator-prey world. We expose the predator models, which are trained in the environment where the birth rate of preys is 0.006 in Fig. 3(a), into a new environment where the birth rate of preys is 0.01. Fig. 3(b) shows that after a period of time for adjustment, the predators adapt to the new environment, and the AI agents as a whole manage to maintain the patterns. Second, we break the binary predator-prey relationships by introducing a second type of prey that does not require group hunting. As shown in Fig. 4, in the case of three species which the LV model may find challenging to analyze, we can still observe the ordered harmonic circles in 3-D space. Third, we investigate the role of individual intelligence by disabling the learning function in the setting of Fig. 3(b). Fig. 5 shows that the AI population does not possess any ordered dynamics anymore if the intelligence of each individual agent is disabled. As such, the whole ecosystem explodes with exponentially-increasing amount of preys and the extinction of predators. The reason why predator goes extinct is that the increased birth rate of preys leads to new distributions on the states, thus the observations; consequently, the original optimal policy of predators becomes suboptimal in the new environment. Given that the number of preys increases exponentially, and the map size is limited, the sheep will soon cover all the blank spaces and the predators can barely aggregate any valid groups for hunting and finally die of starvation.

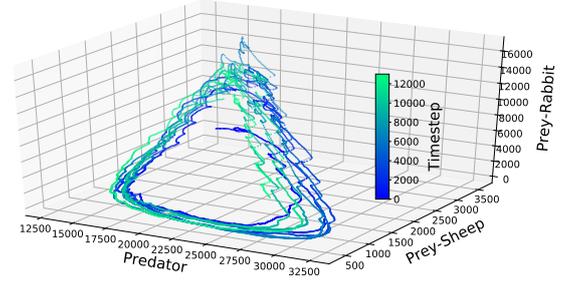

## 5.2 Understanding the Grouping Behaviors

Next, we investigate the dynamics of the collective grouping behaviors. In particular, we intend to find out the relationship between environmental food resources and the proportion of the predators that participate in the group hunting, which we refer to as the "group proportion". In the face of two kinds of preys (one requires group hunting and the other does not), the predators have to make a decision to either join a group for hunting a sheep or hunt a rabbit itself alone. We conduct two experiments with the predator population size equaling 10 thousands and 2 millions, the map size equaling $10^3 \times 10^3$ and $10^4 \times 10^4$ respectively. Acting like a "zookeeper", we supplement the number of preys to a fixed amount if the number drops below a certain threshold. For each supplement, we alternate the types of preys to feed in. Suppose the number of species A is below the threshold, we supply species B. The setting of Q-network is the same as in the study on population dynamics.

AAMAS'18, July 2018, Stockholm, Sweden

Yaodong Yang[†], Lantao Yu[‡], Yiwei Bai[‡], Jun Wang[†], Weinan Zhang[‡], Ying Wen[†], Yong Yu[‡]
University College London[†], Shanghai Jiao Tong University[‡]

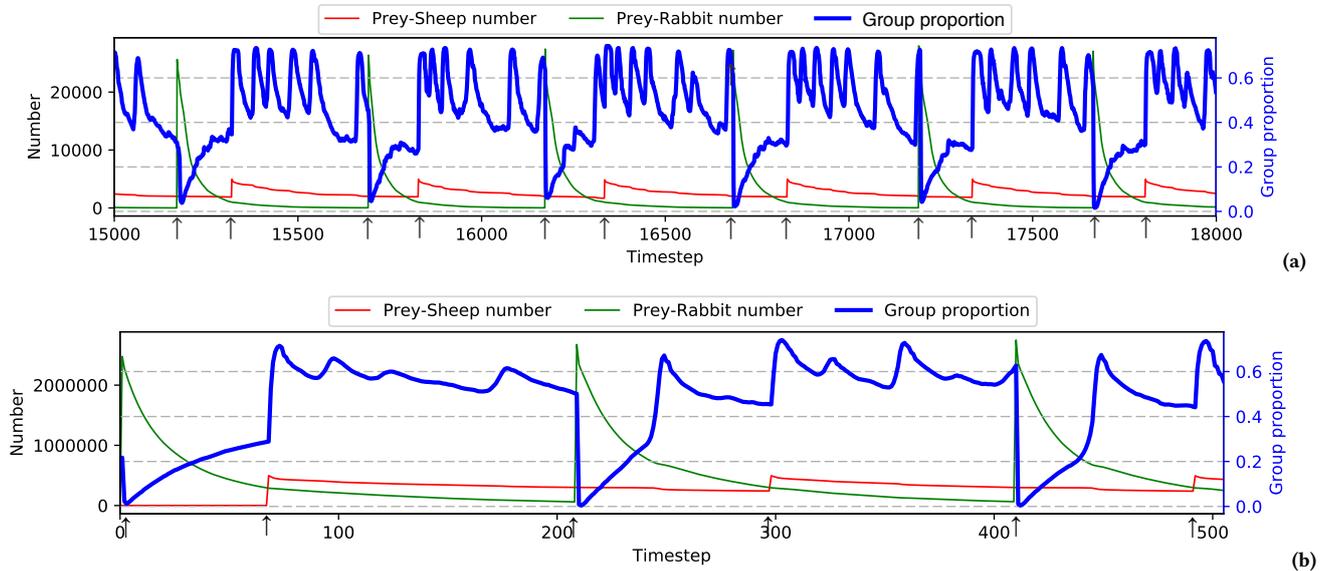

Figure 6: a) Grouping proportion in the predator-prey world where two kinds of preys are fed alternatively. ↑ points out the time step that preys are fed. It tells that when the number of the prey sheep (that requires group hunting) increases, the proportion of groups in AI population increases, and adapting to grouping becomes collective behaviors. Vice verse to the case when the prey rabbit are fed. b) The same experiment on two-million AI population.

As the preys are alternatively fed, the predator's policy needs to react correspondingly to the new environment so as to survive. As shown in Fig. 6(a) and 6(b), the moment right after the rabbits are fed into the environment, the proportion of groups drastically drop down to nearly 0. Predators collectively behave to be selfish rather than to be altruistic to the group. With the number of rabbits being captured, the proportion of grouping behaviors increases mildly again, and meets a spike soon after the sheep are fed into the environment, and reaches another dynamic equilibrium. In a highly-variable environment, the population of predators show the intelligence of adapting their hunting strategies collectively without any external supervisions or controls.

### 5.3 Discussions

Judging from the ordered patterns of the AI population in the predator-prey world, we have reasons to agree with *Lucretius* that a designing intelligence is necessary to create orders in nature. In fact, in understanding the emergence of orders in a system, the theory of *self-organization* proposed in [2] considers that the global ordered dynamics of a system can spontaneously originate from numerous interactions between local individuals that are initially disordered, with no needs of external interventions. The theory predicts the existence of the ordered dynamics from numerous local interactions between the individuals and the system. This could potentially explain the ordered patterns observed on our AI population that has been tested. Meanwhile, according to the theory, the created order is independent of the complexity of the individual involved. For example, the *Lotka-Volterra* dynamics also hold for other natural systems such as the herbivore and the plants, or the parasite and the host. Even though the LV models are based on a set of equations with fixed interaction terms, while our findings depend on intelligent agents driven by consistent learning process, the generalization of the resulting dynamics onto an AI population still leads us to imagine a general law that could unify the artificially created agents with the population we have studied in the natural sciences for long time.

Arguably, in contrast to the *self-organization* theory, reductionist scientists hold a different view that order can only be created by transferring it from external systems. A typical example is *The Second Law of Thermodynamics* [5] stating that the total entropy (the level of disorder) will always increase over time in a *closed* system. Such an idea has widely been accepted, particularly in physics where quantitative analysis is feasible. However, we argue that our findings from the AI population do not go against this law. RL-based agents are not exceptions simply because the environment they "live" in are not *closed*. Whenever a system can exchange matter with its environment, an entropy decrease of that system (orders emerge) is still compatible with the second law. A further discussion on *entropy and life* [41] certainly goes beyond this topic, and we leave it for future work.

## 6 CONCLUSIONS

We conducted an empirical study on an AI population by simulating a predator-prey world where each individual agent was empowered by deep reinforcement learning, and the number of agents is up to millions. We found that the AI population possessed the ordered population dynamics consistent with the *Lotka-Volterra* model in ecology. We also discovered the emergent collective adaptations when the environmental resources changed over time. Importantly, both of the findings could be well explained by the *self-organization* theory from natural sciences.

In the future, we will conduct further experiments on our million-agent RL platform by involving the ideas of leadership, cannibalism, and irrationality for discovering other profound natural principles in the full deep-RL-driven population. In return, we expect our findings could also enlighten an interesting research direction of interpreting the RL-based AI population using the natural science principles developed in the real world, and apply the AI population driven by RL for applications like smart cities or swarm intelligence.

Yaodong Yang[†], Lantao Yu[‡], Yiwei Bai[‡], Jun Wang[†], Weinan Zhang[‡], Ying Wen[†], Yong Yu[‡]
University College London[†], Shanghai Jiao Tong University[‡]